\documentstyle{llncs}

\title{Applying Machine Translation to Two-Stage Cross-Language
Information Retrieval}

\author{\Large Atsushi Fujii and Tetsuya Ishikawa}

\institute{University of Library and Information Science \\ 1-2
Kasuga, Tsukuba, 305-8550, Japan \\ \smallskip {\normalsize\tt
E-mail:~fujii@ulis.ac.jp}}

\newcommand{\eq}[1]{(\ref{#1})}

\newcommand{\shortcite}[1]{\cite{#1}}
\renewcommand{\nocite}[1]{\shortcite{#1}}

\input{psfig.tex}

\begin{document}

\maketitle

\begin{abstract}
  Cross-language information retrieval (CLIR), where queries and
  documents are in different languages, needs a translation of queries
  and/or documents, so as to standardize both of them into a common
  representation. For this purpose, the use of machine translation is
  an effective approach. However, computational cost is prohibitive in
  translating large-scale document collections. To resolve this
  problem, we propose a two-stage CLIR method. First, we translate a
  given query into the document language, and retrieve a limited
  number of foreign documents. Second, we machine translate only those
  documents into the user language, and re-rank them based on the
  translation result. We also show the effectiveness of our method by
  way of experiments using Japanese queries and English technical
  documents.
\end{abstract}

\section{Introduction}
\label{sec:introduction}

The number of machine readable texts accessible via CD-ROMs and the
World Wide Web has been rapidly growing. However, since the content of
each text is usually provided in a limited number of languages, the
notion of information retrieval (IR) has been expanded so that users
can retrieve textual information (i.e., documents) across
languages. One application, commonly termed ``cross-language
information retrieval (CLIR)'', is the retrieval task where the user
presents queries in one language to retrieve documents in another
language. Thus, as can be predicted, CLIR needs to standardize queries
and documents into a common representation, so that monolingual IR
techniques can be applied. From this point of view, existing CLIR can
be classified into three approaches.

The first approach translates queries into the document
language~\cite{ballesteros:sigir-98,davis:sigir-97,fujii:emnlp-vlc-99,nie:sigir-99},
while the second approach translates documents into the query
language~\cite{mccarley:acl-99,oard:amta-98}. The third approach
projects both queries and documents into a language-independent
representation by way of thesaurus
classes~\cite{gonzalo:chum-98,salton:jasis-70} and latent semantic
indexing~\cite{carbonell:ijcai-97,littman:clir-98}.

Although extensive comparative experiments among different approaches
in a rigorous manner are difficult and expensive, a few cases can be
found in past CLIR literature.

Oard~\nocite{oard:amta-98} compared the query and document translation
methods. For the purpose of English-German CLIR experiments, he used
the 21 English queries and SDA/NZZ German collection consisting of
251,840 newswire articles, contained in the TREC-6 CLIR collection.
Then, he showed that the MT-based query translation with the Logos
system was more effective than various types of dictionary-based query
translation methods, and that the MT-based document translation method
further outperformed the MT-based query translation method. Those
findings were salient especially when the length of queries was large.

McCarley~\nocite{mccarley:acl-99} conducted English/French
bidirectional CLIR experiments, where the 141,656 AP English documents
and 212,918 SDA French documents in the TREC-6 and TREC-7 collections
were used, and applied a statistical MT method to both query and
document translation methods. He showed that the relative superiority
between query and document translation methods varied depending on the
source and target language pair. To put it more precisely, in his
case, the quality of French-English translation was better than that
of English-French translation, for both query and document
translations.

In addition, he showed that a hybrid method, where the relevance
degree of each document (i.e., the ``score'') is the mean of those
obtained with query and document translation methods, outperformed
methods based on either query or document translation, irrespective of
the source and target language pair.  Possible rationales include that
since machine translation is not an invertible operation, query and
document translations mutually enhance the possibility that query
terms correspond to appropriate translations in documents.

To sum up, the MT-based document translation approach is potentially
effective in terms of retrieval accuracy.  Besides this, since
retrieved documents are mostly in a user's non-native language, the
document translation approach is significantly effective for browsing
and interactive retrieval.

However, a major drawback of this approach is that the full
translation on large-scale collections is prohibitive in terms of
computational cost. In fact, Oard~\nocite{oard:amta-98}, for example,
spent approximately ten machine-months in translating the SDA/NZZ
collection. This problem is especially crucial in the case where the
number of user languages is large, and documents are frequently
updated as in the Web. Although a fast MT
method~\cite{mccarley:amta-98} was proposed, this method is currently
limited to MT within European languages, which are relatively similar
to one another.

In view of the above discussions, we propose a method to minimize the
computational cost required for the MT-based document translation,
which is fundamentally twofold. First, we translate the query into the
document language, and retrieve a fixed number of top-ranked documents
(one thousand, for example). Second, we machine translate those
documents into the query language, and then re-rank those documents
based on the score, combining those individually obtained with query
and document translation methods. Consequently, it is expected that
the retrieval accuracy is improved with a minimal MT cost.

From a different perspective, our method can be classified as a {\em
two-stage\/} retrieval principle. However, in the monolingual
two-stage IR, the second stage usually involves re-calculation of term
weights and local feedback so as to increase the number of relevant
documents in the final result~\cite{kwok:sigir-98}, and that in the
case of existing two-stage CLIR, multiple stages are used to improve
the quality of query
translation~\cite{ballesteros:sigir-97,davis:sigir-97}.

Section~\ref{sec:system} describes our two-stage CLIR system, where we
elaborate mainly on the MT-based re-ranking method.
Section~\ref{sec:experimentation} then evaluates the performance of
our system, using the NACSIS test collection~\cite{kando:sigir-99},
which consists of 39 Japanese queries and approximately 330,000
technical abstracts in English and Japanese.

\section{System Description}
\label{sec:system}

\subsection{Overview}
\label{subsec:system_overview}

\begin{figure}[htbp]
  \begin{center}
    \leavevmode
    \psfig{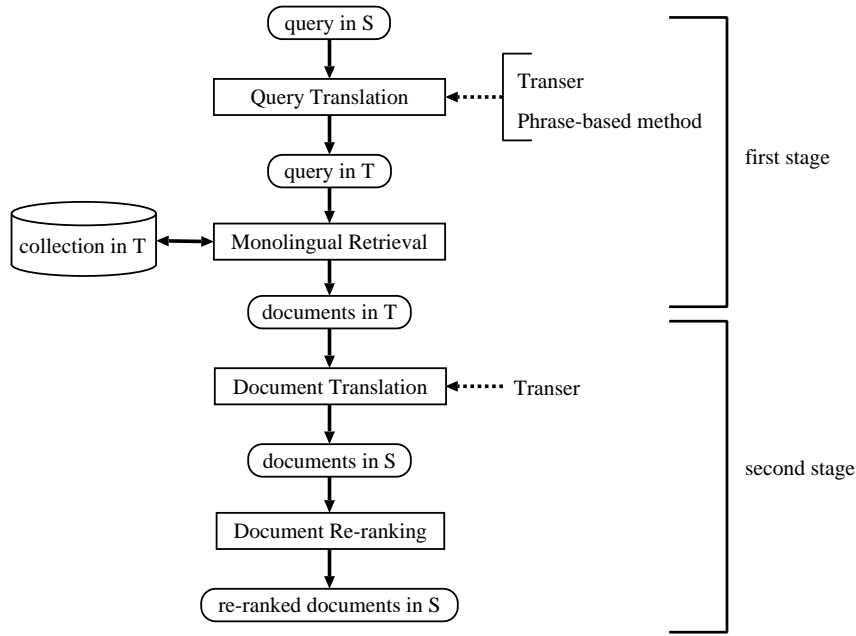}
  \end{center}
  \caption{The overall design of our CLIR system.}
  \label{fig:system}
\end{figure}

Figure~\ref{fig:system} depicts the overall design of our
Japanese/English bidirectional CLIR system, in which we combined query
and document translation modules with a monolingual retrieval
system. In this section, we explain the retrieval process based on
this figure.

First, given a query in the source language (S), a query translation
is performed to output a translation in the target language (T). In
this phase, we use two alternative methods. The first method is the
use of an MT system, for which we use the Transer Japanese/English MT
system.\footnote{Developed by NOVA, Inc.} This MT system uses a
general bilingual dictionary consisting of 230,000 entries, and 19
optional technical dictionaries, among which a computer terminology
dictionary consisting of 100,000 entries is combined with our system.

However, since in most cases, queries consist of a small number of
keywords and phrases, word/phrased-based translation methods are
expected to be comparable with MT systems, in terms of query
translation. Thus, for the second method, we use the Japanese/English
phrase-based translation method proposed by Fujii and
Ishikawa~\cite{fujii:emnlp-vlc-99}, which uses general/technical
dictionaries to derive possible word/phrase translations, and resolves
translation ambiguity based on statistical information obtained from
the target document collection. In addition, for words unlisted in
dictionaries, transliteration is performed to identify phonetic
equivalents in the target language.

Second, the monolingual retrieval system searches a collection for
documents relevant to the translated query, and sorts them according
to the degree of relevance (i.e., the score), in descending order.
For English documents, we use the SMART system~\cite{salton:71}, where
the augmented TF$\cdot$IDF term weighting method (``atc'') is used for
both queries and documents, and the score is computed based on the
similarity between the query and each document in a term vector space.
For Japanese documents, we implemented a retrieval system based on the
vector space model.

Consequently, only the top $N$ documents are selected as an
intermediate retrieval result, where $N$ is a parametric constant.

Third, the top $N$ documents are translated into the source
language. Note that unlike the query translation phase, we use solely
the Transer MT system, because translations are aimed primarily at
human users, and thus the phrase-based translation method potentially
degrades readability of retrieval results.

Finally, the $N$ documents translated are {\em re\/}-ranked according
to the new score. To accomplish this task, we compute the similarity
score between the source query (submitted by the user) and each
translated document in the term vector space, as performed in the
first retrieval stage. We then compute the new score by averaging
those obtained independently with English and Japanese monolingual
similarity computations.  We will elaborate on this process in
Section~\ref{subsec:re-ranking}.

Note that by decreasing the value of $N$, we can decrease the
computational cost required for machine translation. However, this
also decreases the number of relevant documents contained in the top
$N$ set, and potentially dilutes the effectiveness of the re-ranking.
For example, in an extreme case where the top $N$ set contains no
relevant document, the re-ranking procedure does not change the
retrieval accuracy.

The re-ranking procedure is similar to McCarley's hybrid
method~\cite{mccarley:acl-99}, in the sense that his method also
combines scores obtained with query and document
translations. However, unlike McCarley's method, which needs to
translate the entire document collection prior to the retrieval, in
our method the overhead for translating documents is minimized and can
be distributed to each user. In other words, the second stage can be
performed on each client (i.e., users' computers or Web browsers). In
fact, there are a number of commercial Web browsers combined with MT
systems, and thus it is feasible to additionally introduce the
re-ranking function to those browsers.  Besides this, we can easily
replace the MT system with a newer version or those for other language
pairs.

\subsection{MT-based Re-ranking Method}
\label{subsec:re-ranking}

First, given the top $N$ documents retrieved and translated into the
source language, we first compute the similarity score between each
document and the source query provided by the user. Following the
vector space model, both queries and documents are represented by a
vector consisting of statistical factors associated with indexed terms
(i.e., term weights).

In conventional retrieval systems, documents are indexed to produce an
inverted file, prior to the retrieval, so that documents containing
query terms can efficiently be retrieved even from a large-scale
collection.  However, in the case of our re-ranking process, since (a)
the number of target documents is limited, and (b) real-time indexing
degrades the time efficiency, we prefer to use a simple pattern
matching method, instead of the inverted file.

For term weighting, we tentatively use a variation of
TF$\cdot$IDF~\cite{salton:ipm-88,zobel:sigir-forum-98}, as shown in
Equation~\eq{eq:tf_idf}.
\begin{equation}
  \label{eq:tf_idf}
  \begin{array}{lll}
    TF & = & 1 + \log(f_{t,d}) \\
    \noalign{\vskip 1.2ex}
    IDF & = & \log\frac{\textstyle N}{\textstyle n_{t}}
  \end{array}
\end{equation}
Here, $f_{t,d}$ denotes the frequency that term $t$ appears in
document $d$. Note that unlike the common IDF formula, $N$ denotes the
number of documents retrieved in the first stage (see
Section~\ref{subsec:system_overview}), and $n_{t}$ denotes the number
of documents containing term $t$, out of $N$ documents.

One may argue that since in our case where the number of target
documents is considerably smaller than that of the entire collection,
a different term weighting method is needed. For example, the IDF
formula proposed for large-scale document collections may be less
effective for a limited number of documents. However, a preliminary
experiment showed that the use of IDF marginally improved the
performance obtained without IDF. On the other hand, since the
preliminary experiment showed that the use of document length
considerably degraded the performance, we compute the similarity
between the query and each document, as the inner product (instead of
the cosine of the angle) between their associated vectors.

Thereafter, for each document, we combine two similarity scores
obtained in English-English and Japanese-Japanese retrieval processes.
We shall call them $ESIM$ and $JSIM$, respectively.  Since those two
similarity scores have different ranges, we use a geometric mean,
instead of an arithmetic mean, as shown in
Equation~\eq{eq:new_similarity}.
\begin{eqnarray}
  \label{eq:new_similarity}  
  SIM & = & ESIM^\alpha \cdot JSIM^\beta
\end{eqnarray}
Here, $SIM$ is the final similarity score with which we re-rank the
top $N$ documents, and $\alpha$ and $\beta$ are parametric constants
used to control the degree to which $ESIM$ and $JSIM$ affect the
computation of $SIM$. However, in the case where either $ESIM$ or
$JSIM$ is zero, the value of $SIM$ always becomes zero, disregarding
the value of the other similarity score. To avoid this problem, in
such a case we arbitrarily assign the value 0.0001 to either $ESIM$ or
$JSIM$ that takes zero.

Possible factors to set values of $\alpha$ and $\beta$ include the
quality of Japanese-English and English-Japanese translations. In the
case where the quality of one of the translations is considerably
lower, $\alpha$ and $\beta$ must be properly set so as to decrease the
effect of the similarity score through the lower quality
translation. Generally speaking, the quality of English-Japanese
translation is higher than that of Japanese-English translation,
because morphological and syntactic analyses for Japanese are usually
more crucial than those for English.  However, we empirically set
\mbox{$\alpha=\beta=1$}, that is, we consider $ESIM$ and $JSIM$
equally in the re-ranking process.

\medskip
\section{Experimentation}
\label{sec:experimentation}

\subsection{Methodology}
\label{subsec:eval_overview}

We investigated the performance of several versions of our system in
terms of Japanese-English CLIR, where each system outputs the top
1,000 documents, and the TREC evaluation software was used
to calculate non-interpolated average precision values.

For the purpose of our experiments, we used the official version
of the NACSIS test collection~\cite{kando:sigir-99}. This collection
consists of 39 Japanese queries and approximately 330,000 documents
(in either a combination of English and Japanese or either of the
languages individually), collected from technical papers published by
65 Japanese associations for various fields.

Each document consists of the document ID, title, name(s) of
author(s), name/date of conference, hosting organization, abstract and
keywords, from which titles, abstracts and keywords were indexed by
the SMART system. We used as target documents 187,081 entries that are
in both English and Japanese.

Each query consists of the query ID, title of the topic, description,
narrative and list of synonyms, from which we used only the
description.  Figure~\ref{fig:query} shows example descriptions
(translated into English by one of the authors).

The NACSIS collection was produced for a TREC-type (CL)IR workshop
held by NACSIS (National Center for Science Information Systems,
Japan) in 1999.\footnote{See {\tt
http://www.rd.nacsis.ac.jp/\~{}ntcadm/workshop/work-en.html} for
details of the NACSIS workshop.} In this workshop, each participant
was allowed to submit more than one retrieval result using different
methods. However, at least one result had to be gained with only the
description field in queries. According to experimental results
reported in the proceedings of the workshop~\cite{ntcir-99}, in the
case where only the description field was used, average precision
values ranged from 0.021 to 0.182.

Relevance assessment was performed based on the pooling
method~\cite{voorhees:sigir-98}. To put it more precisely, candidates
for relevant documents were first pooled by multiple retrieval systems
(primarily systems that participated in the NACSIS
workshop). Thereafter, for each candidate document, human expert(s)
assigned one of three ranks of relevance, that is, ``relevant'',
``partially relevant'' and \mbox{``irrelevant''.} The average number
of candidate documents pooled for each query is 2,509, among which the
number of relevant and partially relevant documents are approximately
21 and 6, respectively.  In our experiments, we did not regard
``partially relevant'' documents as relevant ones, because
interpretation of ``partially relevant'' is not fully clear to the
authors. Note that since the NACSIS collection does not contain
English queries, we cannot estimate a baseline for Japanese-English
CLIR performance using English-English IR.

In the following two sections, we will show experimental results in
terms of the first and second stages (i.e., query translation methods
and the MT-based re-ranking method), respectively.

\begin{figure}[htbp]
  \begin{center}
    \leavevmode
    \small
    \begin{tabular}{cl} \hline\hline
      ID & {\hfill\centering Description\hfill} \\ \hline
      0032 & middleware construction in network collaboration \\
      0035 & digital libraries in distributed systems \\
      0036 & problems related to groupwares in mobile communication \\ 
      0062 & life-long education and volunteer \\
      0065 & image retrieval based on genetic algorithm \\
      \hline
    \end{tabular}
    \caption{Example query descriptions in the NACSIS collection.}
    \label{fig:query}
  \end{center}
\end{figure}

\medskip
\subsection{Evaluation of Query Translation Methods}
\label{subsec:eval_query_translation}

The primal objective in this section is to compare the effectiveness
of the phrase-based translation method proposed by Fujii and
Ishikawa~\nocite{fujii:emnlp-vlc-99} and one based on the Transer MT
system, in terms of Japanese-English query translation. While the
former method is aimed solely at words and phrases, the MT system can
also be used for full sentences. In addition, since both methods are,
to some extent, complementary to each other, we theoretically gain a
query expansion effect, combining query terms translated by individual
methods. In view of those above factors, we compared the following
query translation methods:
\begin{itemize}
\item the use of the Transer MT system for full sentences contained in 
  the description field (``MTS''),
\item the use of the Transer MT system for content words and phrases
  extracted from the description field, for which the ChaSen
  morphological analyzer~\cite{matsumoto:chasen-97} was used
  (``MTP''),
\item the phrase-based translation method applied to the same words
  and phrases as used for the MTP method (``PBT''),
\item the use of query terms obtained with both MTP and PBT, where
  terms outputed by both methods are considered to appear twice in the
  query (``MPBT'').
\end{itemize}
Table~\ref{tab:avg_pre} shows the non-interpolated average
precision values, averaged over the 39 queries, for different query
translation methods listed above.  The second column denotes the
average number of query terms provided with each translation method,
some of which were potentially discarded as stopwords by the SMART
system. The third column denotes average precision values for
different query translation methods. We will explain the fourth and
fifth columns in Section~\ref{subsec:eval_re-ranking}.

Looking at this table, one can see that while two MT-based methods,
that is, MTS and MTP, were quite comparable in performance, and that
PBT outperformed both of them. In the case of PBT, the transliteration
successfully identified English equivalents for {\it katakana\/} words
unlisted in the word dictionary, such as ``{\it
coraboreishon\/}~(collaboration)'' and ``{\it mobairu\/}~(mobile)'',
which the MT-based methods failed to translate.  Another reason was
due to the difference in dictionaries used.  Generally speaking, PBT
tended to output technical words more than the MT-based methods. For
example, for Japanese phrases ``{\it fukusuu-deeta\/}'' and ``{\it
sekitsui-doubutsu}'', PBT outputed ``multiple data'' and ``craniate'',
while MTS/MTP outputed ``more than one data'' and ``vertebrate'',
respectively. Note that this effect was evident partially because the
NACSIS collection consists of technical documents. In addition, MPBT
further improved the performance of PBT. Although the difference
between PBT and MPBT was marginal, it is worth utilizing both the
MT-based and phrase-based methods, if available, for query
translation.

\begin{table}[htbp]
  \begin{center}
    \caption{Non-interpolated average precision values,
    averaged over the 39 queries.}
    \medskip
    \leavevmode
    \small
    \tabcolsep=3pt
    \begin{tabular}{lrcll} \hline\hline
      Query Translation & & &
      \multicolumn{2}{c}{Avg. Precision with Re-ranking} \\
      \cline{4-5}
      Method & \# of Terms &
      Avg. Precision & {\hfill\centering MT\hfill} & {\hfill\centering
      HT\hfill} \\ \hline
      ~~~~~~~~MTS  & 16.6~~~~ & 0.1124 & 0.1770 (+57.5\%) & 0.2297
      (+104.3\%) \\
      ~~~~~~~~MTP  & 8.7~~~~ & 0.1134 & 0.1746 (+54.0\%) & 0.2217
      (+95.5\%) \\
      ~~~~~~~~PBT  & 6.1~~~~ & 0.1403 & 0.2013 (+43.5\%) & 0.2295
      (+63.6\%) \\
      ~~~~~~~~MPBT & 13.1~~~~ & 0.1426 & 0.1986 (+39.3\%) & 0.2356
      (+65.2\%) \\
      \hline
    \end{tabular}
    \label{tab:avg_pre}
  \end{center}
\end{table}

To validate those above results in a thorough manner, we used the
non-parametric Wilcoxon matched-pairs signed-test for statistical
testing (at the 5\% level), which investigates whether the difference
in average precision is meaningful or simply due to
chance~\cite{hull:sigir-93,keen:ipm-92,srinivasan:ipm-90}. We found
that differences in average precision values for pairs ``MTP versus
MTS'', ``MPBT versus MTS'', and ``MPBT versus MTP'' were significant,
although for other pairs, we could not obtain sufficient evidence to
conclude a statistical significance. To sum up, we concluded that in
query translation, a combination of MT-based and phrase-based
translation methods was more effective than a method relying solely on
the MT system.

\medskip
\subsection{Evaluation of the MT-based Re-ranking Method}
\label{subsec:eval_re-ranking}

First, we consider Table~\ref{tab:avg_pre} again, where the fourth
column ``MT'' denotes the average precision values for each query
translation method, combined with the MT-based re-ranking
method. Throughout our experimentation in this paper, the best average
precision value by an automatic method was 0.2013 (i.e., one obtained
by PBT combined with the MT-based re-ranking method), which is
relatively high, when compared with average precision values reported
in the NACSIS workshop (ranging from 0.021 to 0.182).

For each query translation method, the improvement in average
precision from one without the re-ranking, which is generally
noticeable, is indicated in parentheses.  In fact, we used the
Wilcoxon test again, as conducted in
Section~\ref{subsec:eval_query_translation}, and confirmed that every
improvement was statistically significant. To sum up, the MT-based
re-ranking method we proposed was generally effective, irrespective of
the query translation method combined, in terms of CLIR performance.

Second, we conducted an error analysis for queries for which the
re-ranking method degraded the average precision, and found that
roughly two thirds of errors were due to ambiguity in the document
translation.  For example, the English word ``library'' was often
incorrectly translated into ``{\em raiburari\/}~(library as a
software)'', whereas the original query was intended to ``{\em
toshokan\/}~(library as an institution)''.

Third, to estimate the upper bound of the re-ranking method, as
denoted in the fifth column ``HT'', we used as human translations
Japanese documents comparable to English ones in the NACSIS
collection. By comparing the results of ``MT'' and ``HT'', one can see
that MT systems with a higher quality, if available, are expected to
further improve our CLIR system. In fact, when we manually corrected
inappropriate translations in translated documents, such as
``library~({\em raiburari/toshokan\/})'' above, the average precision
of ``MT'' became almost equivalent to that of ``HT''.

Noted that when combined with the re-ranking method, differences among
query translation methods in average precision were relatively
overshadowed.  In the case of ``MT'', the Wilcoxon test showed that
differences in only pairs ``MPBT versus MTS'' and ``MPBT versus MTP''
were significant, while in the case of ``HT'', none of the differences
were identified as significant.

Fourth, we investigated how the number of documents retrieved in the
first stage (i.e., the value of $N$ in Section~\ref{sec:system})
affected the performance of the re-ranking method. As discussed in
Section~\ref{subsec:system_overview}, in real world usage, one has to
consider the trade-off between the retrieval accuracy (i.e., average
precision in our case) and overhead required for the document
translation.

Table~\ref{tab:docnum_avgpre} shows the results, where average
precision values in the column \mbox{``1,000''} correspond to those in
Table~\ref{tab:avg_pre}. By comparing average precision values for
each of four query translation methods (i.e., MTS, MTP, PBT and MPBT)
and those suffixed with ``+MT'' and ``+HT'' in
Table~\ref{tab:docnum_avgpre}, one can see that the re-ranking methods
were effective, irrespective of the number of documents retrieved.  In
other words, it is expected that we can minimize the overhead in
translating documents, without decreasing the retrieval accuracy.

Table~\ref{tab:xtime} shows CPU time (sec.) required for the document
translation and re-ranking procedures, averaged over four different
query translation methods. In the case of \mbox{$N=1,000$}, the total
CPU time was approximately three minutes, which is perhaps not
tolerable for a real-time usage. However, for small values of $N$
(e.g., 50 and 100), the CPU time was more acceptable and practical,
maintaining the improvement of retrieval accuracy.

\begin{table}[htbp]
  \begin{center}
    \caption{The relation between the number of documents retrieved in 
    the first stage and non-interpolated average precision
    values, averaged over the 39 queries.}
    \medskip
    \leavevmode
    \small
    \tabcolsep=4pt
    \begin{tabular}{lccccccc} \hline\hline
      & \multicolumn{7}{c}{\# of Documents Retrieved ($N$)} \\
      \cline{2-8}
      {\hfill\centering Method\hfill} & 50 & 100 & 200 & 400 & 600 &
      800 & 1,000 \\ \hline
      MTS & 0.0949 & 0.1017 & 0.1074 & 0.1101 & 0.1112 & 0.1119 &
      0.1124 \\
      MTS+MT & 0.1341 & 0.1556 & 0.1673 & 0.1698 & 0.1720 & 0.1736 &
      0.1770 \\
      MTS+HT & 0.1666 & 0.1901 & 0.2070 & 0.2173 & 0.2230 & 0.2259 &
      0.2297 \\
      \hline
      MTP & 0.0953 & 0.1020 & 0.1085 & 0.1113 & 0.1123 & 0.1131 &
      0.1134 \\
      MTP+MT & 0.1449 & 0.1584 & 0.1692 & 0.1711 & 0.1728 & 0.1750 &
      0.1746 \\
      MTP+HT & 0.1619 & 0.1819 & 0.2017 & 0.2105 & 0.2165 & 0.2203 &
      0.2217 \\
      \hline
      PBT & 0.1215 & 0.1301 & 0.1355 & 0.1385 & 0.1394 & 0.1399 &
      0.1403 \\
      PBT+MT & 0.1553 & 0.1723 & 0.1866 & 0.1954 & 0.1978 & 0.2005 &
      0.2013 \\
      PBT+HT & 0.1722 & 0.1915 & 0.2097 & 0.2212 & 0.2241 & 0.2279 &
      0.2295 \\
      \hline
      MPBT & 0.1229 & 0.1305 & 0.1376 & 0.1405 & 0.1416 & 0.1421 &
      0.1426 \\
      MPBT+MT & 0.1690 & 0.1766 & 0.1901 & 0.1946 & 0.1958 & 0.1967 &
      0.1986 \\
      MPBT+HT & 0.1814 & 0.1968 & 0.2142 & 0.2242 & 0.2301 & 0.2319 &
      0.2356 \\
      \hline
    \end{tabular}
    \label{tab:docnum_avgpre}
  \end{center}
\end{table}

\begin{table}[htbp]
  \begin{center}
    \caption{CPU time for document translation and re-ranking (sec.).}
    \medskip
    \leavevmode
    \small
    \tabcolsep=4pt
    \begin{tabular}{lrrrrrrr} \hline\hline
      & \multicolumn{7}{c}{\# of Documents Retrieved ($N$)} \\
      \cline{2-8}
      & {\hfill\centering 50 \hfill} &
      {\hfill\centering 100 \hfill} &
      {\hfill\centering 200 \hfill} &
      {\hfill\centering 400 \hfill} &
      {\hfill\centering 600 \hfill} &
      {\hfill\centering 800 \hfill} &
      {\hfill\centering 1,000 \hfill}
      \\ \hline
      translation & 9.5 & 17.7 & 33.3 & 65.6 & 106.2 & 139.3 & 175.1 \\
      re-ranking  & 0.2 &  0.3 &  0.6 &  1.2 &   1.8 &   2.4 &   3.0 \\ 
      total       & 9.7 & 18.0 & 33.9 & 66.8 & 108.0 & 141.7 & 178.1 \\
      \hline
      \multicolumn{8}{r}{(Pentium III 700MHz)}
    \end{tabular}
    \label{tab:xtime}
  \end{center}
\end{table}

\section{Conclusion}
\label{sec:conclusion}

Reflecting the rapid growth in utilization of machine readable texts,
cross-language information retrieval (CLIR) has variously been
explored in order to facilitate retrieving information across
languages.

In brief, existing CLIR systems are classified into three approaches:
(a) translating queries into the document language, (b) translating
documents into the query language, and (c) representing both queries
and documents in a language-independent space. Among these approaches,
the second approach, based on machine translation, is effective in
terms of retrieval accuracy and user interaction.  However, the
computational cost in translating large-scale document collections is
prohibitive.

To resolve this problem, we proposed a two-stage CLIR method, in which
we first used a query translation method to retrieve a fixed number of
documents, and then applied machine translation only to those
documents, instead of the entire collection, to improve the document
ranking.

Through Japanese-English CLIR experiments using the NACSIS collection,
we showed that our two-stage method significantly improved average
precision values obtained solely with query translation methods. We
also showed that our method performed reasonably, even in the case
where the number of retrieved documents was relatively small.

\section*{Acknowledgments}

The authors would like to thank NOVA, Inc. for their support with the
Transer MT system, and Noriko Kando (National Institute of
Informatics, Japan) for her support with the NACSIS collection.

\bibliographystyle{jplain}

\end{document}